# CoMoE: Contrastive Representation for Mixture-of-Experts in Parameter-Efficient Fine-tuning

**Jinyuan Feng**[1,2*] **Chaopeng Wei**[3*] **Tenghai Qiu**[1†] **Tianyi Hu**[1,2] **Zhiqiang Pu**[1,2]
[1] Institute of Automation, Chinese Academy of Sciences
[2] School of Artificial Intelligence, University of Chinese Academy of Sciences
[3] University of Science and Technology Beijing

## Abstract

In parameter-efficient fine-tuning, mixture-of-experts (MoE), which involves specializing functionalities into different experts and sparsely activating them appropriately, has been widely adopted as a promising approach to trade-off between model capacity and computation overhead. However, current MoE variants fall short on heterogeneous datasets, ignoring the fact that experts may learn similar knowledge, resulting in the underutilization of MoE's capacity. In this paper, we propose Contrastive Representation for MoE (CoMoE), a novel method to promote modularization and specialization in MoE, where the experts are trained along with a contrastive objective by sampling from activated and inactivated experts in top-k routing. We demonstrate that such a contrastive objective recovers the mutual-information gap between inputs and the two types of experts. Experiments on several benchmarks and in multi-task settings demonstrate that CoMoE can consistently enhance MoE's capacity and promote modularization among the experts.

## 1 Introduction

Parameter-Efficient Fine-Tuning (PEFT) has emerged to efficiently adapt Large Language Models (LLMs) to downstream tasks by updating only a subset of parameters, significantly reducing computational and memory overhead (Hu et al., 2022a; Liu et al., 2022; He et al., 2021). However, it struggles with substantially increased dataset sizes, especially heterogeneous training datasets, which poses a significant practical challenge (Huang et al., 2024; Wang et al., 2024b). Mixture-of-Experts (MoE) offers a versatile solution to the challenge for its modular design (Zhang et al., 2024).

Thus, Low-rank Adaptation (LoRA), as a popular and effective PEFT method, has been widely integrated with MoE (Dou et al., 2024; Li et al., 2024), leveraging MoE's modularity to enhance the model's capacity and performance. By sparsely activating a subset of experts, LoRA's MoE variants achieve efficient training on heterogeneous datasets and allocate the experts adaptively (Tian et al., 2024). Specifically, the sparse activation is controlled through a router mechanism (e.g., top-k routing) that dispatches inputs to the activated experts. Basically, given an input token, only a subset of specialized experts contribute to the output, while other irrelevant experts remain inactive.

Ideally, each expert should specialize in distinct representation subspaces and semantic skills, thereby collaboratively enhancing the model's representational capacity and enabling a broader spectrum of knowledge (Liu et al., 2023). However, despite the explicit division into multiple experts in MoE architecture, its modularization degree remains questionable. Two issues persist: (1) **expert knowledge redundancy**, where insufficient specialization constraints lead to overlapping functionalities among experts, limiting the model's capacity (Feng et al., 2025); (2) **expert load imbalance**, where inadequate modularity and specialization during training result in frequent activation of only a subset of experts, which underutilizes other experts and contradicts its original design intent. Consequently, as some studies have indicated (Qian et al., 2024), simply stacking more experts does not linearly improve performance; instead, it leads to a performance bottleneck. Existing studies propose load balance loss (Li et al., 2024) and localized balancing constraint (Dou et al., 2024) to alleviate the mentioned issues, but that is still far from enough.

In this paper, we propose a novel perspective to promote the specialization of experts. As illustrated in Fig. 1, building upon top-k routing, we categorize the experts into activated experts and inactivated experts. Then, we quantify the specialization of experts by mutual information (MI) between the input token and the two types of experts. To promote expert specialization, we define an MI gap, which is derived from the aforementioned MI, and aim to maximize it. In practice, based on the InfoNCE theory (Oord et al., 2018), such an MI gap can be approximated via a contrastive objective by using positive samples from the activated experts and negative samples from the inactive experts (Lan et al., 2024; Wen et al., 2024). The contrastive objective is incorporated as an auxiliary objective during training, encouraging specialization and modularization among experts. We name the proposed method **Co**ntrastive Representation for MoE (CoMoE), a novel MoE variant. Empirically, we evaluate CoMoE on diverse benchmarks, showcasing its remarkable performance on heterogeneous tasks. Summary of our contributions:

- We define an MI gap to quantify expert specialization and redundancy in top-k routing, with contrastive learning providing an efficient estimation approach.
- We propose a novel MoE variant, named CoMoE, which incorporates an auxiliary contrastive objective to enhance expert specialization and modularization.
- Comprehensive experiments are conducted to demonstrate that our method consistently improves MoE on heterogeneous tasks.

*Equal contributions.
†Corresponding author.

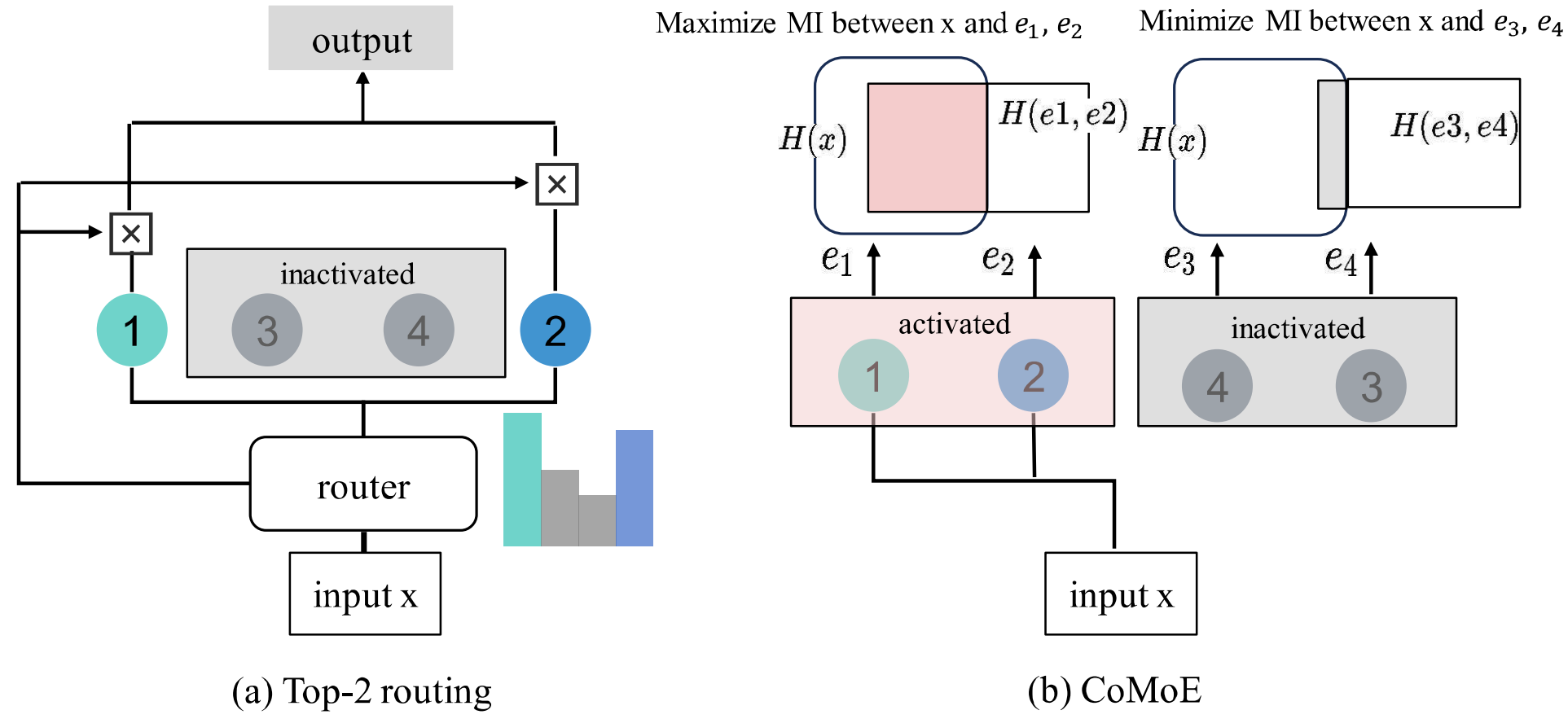

(a) Top-2 routing

(b) CoMoE

Figure 1: Given an input token $x$, (a) illustrates a workflow of top-2 routing, which serves as a fundamental mechanism of CoMoE; (b) illustrates the motivation of CoMoE: maximizing MI between input $x$ and activated experts while minimizing MI between input $x$ and inactive experts.

## 2 Preliminaries

**LoRA Basics** LoRA (Hu et al., 2022a) introduces a pair of low-rank matrices $\mathbf{A}$ and $\mathbf{B}$ to reparameterize the pretrained weights $\mathbf{W_0}$ in a residual manner. Specifically, input $x$ is processed through both the frozen weights and the low-rank matrices:

$$y' = \mathbf{W_0}x + \mathbf{BA}x, \tag{1}$$

where $y'$ denotes the output, with $\mathbf{A} \in \mathbb{R}^{r \times d_2}$ and $\mathbf{B} \in \mathbb{R}^{d_1 \times r}$. The rank $r \ll \min(d_1, d_2)$ is significantly small to reduce tunable parameters.

**Mixture of Experts** In LoRA's MoE variants, the original LoRA module is substituted with $n$ parallel experts, each denoted as $\{E_i(x) = \mathbf{B_i A_i}x\}_{i=1}^n$. These experts are activated via a router $g(x; \mathbf{G})$ to process the input collaboratively. Specifically, given an input $x$, the router calculates the importance of each expert, and the output $y'$ is computed residually as a weighted sum of outputs from the experts:

$$y' = \mathbf{W_0}x + \sum_{i=1}^{n} g_i(x; \mathbf{G}) E_i(x), \tag{2}$$

where $g_i(x; \mathbf{G})$ represents the weight of the $i$-th expert, and $E_i(x)$ denotes the output of expert $i$.

**Top-k Routing** Top-k routing is a common and effective routing strategy of the router $g(x; \mathbf{G})$ in MoE, which sparsely activates a subset of the experts. Specifically, only the top $k$ experts with the highest values in $g(x; \mathbf{G})$ are activated. Then, $g(x; \mathbf{G})$ is renormalized for the activated experts.

The renormalization is computed as follows:

$$\hat{g}_i(x) = \begin{cases} \frac{g_i(x)}{\sum_{j \in \text{top}(g(x),k)} g_j(x)} & \text{if } i \in \text{top}(g(x), k) \\ 0 & \text{if } i \notin \text{top}(g(x), k) \end{cases}, \quad (3)$$

where $\text{top}(g(x), k)$ returns the indices of the largest $k$ elements in $g(x)$.

## 3 Related Works

### 3.1 Parameter-Efficient Fine-Tuning

PEFT methods can be categorized into Adapter Tuning, Prompt Tuning, and Prefix Tuning: **Adapter Tuning** (Hu et al., 2023, 2022a; Zhang et al., 2023b) includes algorithms like BitFit (Ben-Zaken et al., 2022), which links learnable vectors to hidden states using multiplication or addition. (IA)$^3$ (Liu et al., 2022) introduces interpretable adapters to enhance model transparency and task-specific adaptability. **Prompt-Tuning** (Li and Liang, 2021; Liu et al., 2021; Zhu et al., 2024) introduces learnable prefixes to the input sequence. **Prefix-Tuning** (Li and Liang, 2021) modifies hidden states by adding embeddings, making it efficient for few-shot tasks.

LoRA (Hu et al., 2022a) reduces the number of tunable parameters by introducing low-rank matrices into pre-trained weights, enhancing performance and efficiency. Currently, LoRA becomes the most popular and commonly used PEFT method (Cui et al., 2023; Zhang et al., 2022). Numerous studies have focused on improving LoRA: Tied-LoRA (Renduchintala et al., 2024) further reduces trainable parameters by applying weight binding. AdaLoRA (Zhang et al., 2023a) utilizes singular value decomposition to decompose the weight matrices and eliminates insignificant singular values, thereby simplifying the update process. LoraHub (Huang et al., 2023) employs dynamic LoRA composition with gradient-free weight optimization. MoDULA (Ma et al., 2024) uses a mixture of domain-specific and universal LoRA with dynamic weight balancing. DoRA (Liu et al., 2024) decomposes weights into magnitude and direction components for efficiency.

### 3.2 Mixture of Experts

Recent research has increased its focus on combining LoRA with MoE (Shazeer et al., 2017; Jacobs et al., 1991) for adaptable and scalable LLMs, emerging as a balance between model capacity and computation overhead (Li et al., 2024; Zhang et al., 2024). Generally, MoE is employed for two primary purposes: (1)**Avoiding catastrophic forgetting**: LoRAMoE (Dou et al., 2024) leverages token-based routing to mitigate knowledge loss. MOELoRA (Liu et al., 2023) enhances expert selection in multi-tasking. MoRAL (Yang et al., 2024) adapts to new tasks in lifelong learning while preserving old knowledge. (2) **Model efficiency**: HydraLoRA (Tian et al., 2024) adopts an asymmetric LoRA structure without domain expertise, and MiLoRA (Zhang et al., 2024) employs a prompt-aware routing to reduce latency. LoRAMoE (Dou et al., 2024) incorporates a localized balancing constraint to achieve balanced workloads. SCMoE (Shi et al., 2024) leverages unchosen experts to enhance parameter efficiency. MoLoRA (Zadouri et al., 2023) achieves highly parameter-efficient MoE through sparse expert activation, low-rank decomposition, and parameter sharing. C-Poly (Wang et al., 2024a) enables customizable integration of heterogeneous parameter-efficient modules through meta-learned strategies. Current methods focus on architectural expert partitioning but neglect capacity underutilization caused by expert redundancy.

## 4 Methods

In this section, we first define an MI gap for top-k routing. Then we derive a contrastive objective to estimate the gap with learning expert representations. Finally, we present a training approach that incorporates the contrastive objective as an auxiliary loss.

### 4.1 Motivation

As demonstrated in OMoE (Feng et al., 2025), the vanilla MoE variant lacks specialization and modularity, causing LoRA experts to collapse into similar distributions. The lack of specialization and redundant knowledge minimizes the utilization of capacity, which exacerbates performance degradation in heterogeneous tasks. Existing MoE variants (Li et al., 2024; Liu et al., 2023; Luo et al., 2024) leverage balance loss to promote specialization among the experts, but fall far short.

Ideally, experts should exhibit modularity and high specialization with minimal redundancy. To quantify these properties, we leverage MI, a basic concept in information theory, to evaluate the dependence between inputs and experts in MoE. An intuitive illustration of expert specialization is provided in Appendix K. Thus, we formalize the aforementioned idea using MI to quantify the specialization and redundancy between the experts in top-k routing:

- **Maximizing the MI between inputs and activated experts**: Promote activated experts to respond to inputs that highly match, thereby encouraging specialization. In addition, MI can serve as an information bottleneck that filters irrelevant noise.
- **Minimizing the MI between inputs and inactivated experts**: Suppress the response of inactivated experts to irrelevant inputs, thereby preventing multiple experts from learning similar representations.

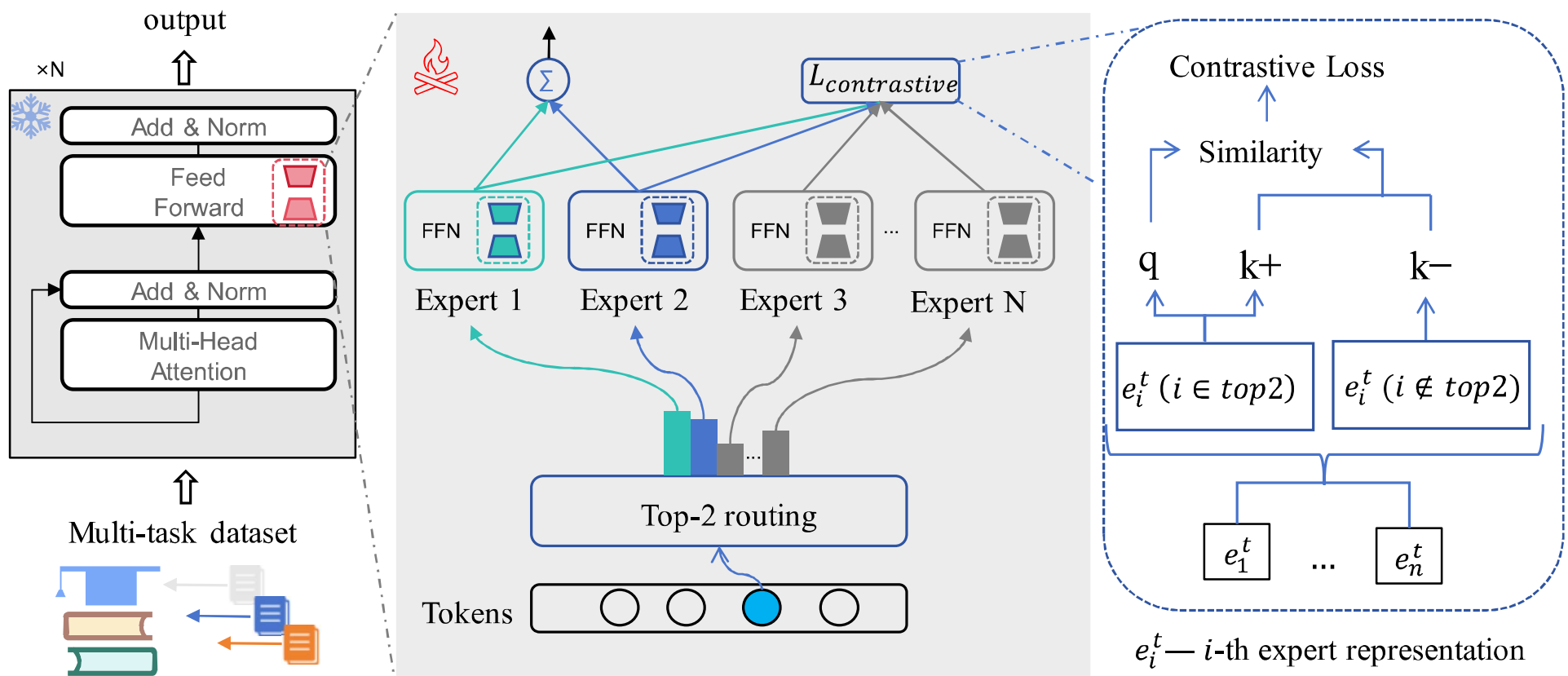

Figure 2: Architecture of CoMoE. Activated experts $e_i^t$, $i \in$ top2 in top-2 routing are selected as the query $q$ and positive keys $k^+$, while inactivated experts $e_i^t$, $i \notin$ top2 serve as negative keys $k^-$.

### 4.2 MI Gap for Top-k Routing

Building upon the motivation, we begin by defining an MI Gap for input token $x$ and task experts set $M$ in top-k routing. For simplicity, we denote $I(\cdot;\cdot)$ as MI, $H(\cdot)$ as Shannon entropy, and $E_i(\cdot)$ as the network of expert $i$. Correspondingly, the representations of activated and inactivated experts are formulated as $e^+ = E^+(x)$ and $e^- = E^-(x)$.

Given an input token, an ideal expert in MoE should retain specialized knowledge while discarding redundant knowledge with others. Intuitively, this equals the uncertainty reduction in the activated experts given the input token. We use $p$ to denote the joint distribution as well as their associated marginals. Then the MI between the input token $x$ and task experts set $M$ is defined as:

$$I\left(x;M\right) = \mathbb{E}_{x,\mathcal{M}\sim\mathcal{D}}\left[\log\frac{p(M|x)}{p(M)}\right] \tag{4}$$

where $p(M|x)$ and $p(M)$ follow empirical distributions according to the existing input and output dataset $\mathcal{D}$. Then, we establish a lower bound for $I\left(x;M\right)$ to convert it to the MI between $x$ and the expert representation $e = E(x)$:

$$\begin{aligned} I(x;M) &= \mathbb{E}_{x,\mathcal{M}}\left[\log\frac{p(x\mid M)}{p(x)}\right] \\ &= \mathbb{E}_{x,\mathcal{M}}\left[\log\int_e\frac{p(x\mid e)p(e\mid M)}{p(x)}de\right] \\ &= \mathbb{E}_{x,\mathcal{M}}\left[\log\mathbb{E}_e\frac{p(x\mid e)}{p(x)}\right] \\ &= \mathbb{E}_{x,\mathcal{M}}\left[\log\mathbb{E}_e\frac{p(e\mid x)}{p(e)}\right], \end{aligned} \tag{5}$$

where $p(e \mid x)$ can be approximated by the expert network $E(x)$. Since $\log(\cdot)$ is a convex function, Jensen's inequality yields the theoretical lower bound of $I\left(x;M\right)$:

$$I(x;M) \geq \mathbb{E}_{x,e,\mathcal{M}}\left[\log\frac{p(e\mid x)}{p(e)}\right]. \tag{6}$$

In the top-k routing, the input token $x$ and the expert representation $e$ form samples $(x,e)$, which can be categorized into two different datasets: $(x,e^+) \sim \mathcal{D}_{\text{top-k}}$, samples from the activated experts; and $(x,e^-) \sim \mathcal{D}_{\neg\text{top-k}}$, samples from the inactivated experts. For clarity, we denote $M^+$ as the activated task experts and $M^-$ as the inactivated. Then, we define the MI gap for top-k routing as:

**Definition 1** *MI Gap for activated experts and inactivated experts in top-k routing is defined by:*

$$\Delta I = I_{top\text{-}k}\left(x, M^+\right) - I_{\neg top\text{-}k}(x, M^-), \tag{7}$$

*where $I_{top\text{-}k}$ is the MI between input token $x$ and activated experts $M^+$, and $I_{\neg top\text{-}k}$ is the MI between input token $x$ and inactivated experts $M^-$.*

The MI terms $I_{\text{top-k}}$ and $I_{\neg\text{top-k}}$ can be converted into a similar form as Eq. 5. To maximize the MI gap $\Delta I$, it is necessary to maximize the MI between $x$ and the activated task experts $M^+$ while minimizing the MI between $x$ and $M^-$. When the experts exhibit high specialization, each expert only yields high mutual information for specific token subsets. Concurrently, when knowledge redundancy among the experts is minimized, the mutual information between tokens and inactivated experts approaches zero.

### 4.3 Contrastive Representation for MI Gap

To estimate the MI gap, Eq. 6 provides a lower bound for the MI between input token $x$ and task experts $M$. In practice, $p(e|x)/p(e)$ cannot be calculated directly, thus we use two score functions $h_1(\cdot)$ and $h_2(\cdot)$ to measure the information density ratio, which preserves the MI between $x$ and $e$ for the activated and inactivated experts, respectively. Let $h_1(x, e^+) \propto p(e^+|x)/p(e^+)$ and $h_2(x, e^-) \propto p(e^-|x)/p(e^-)$. Then, we adopt contrastive learning to estimate the MI gap.

Instead of estimating $I_{\text{top-k}}$ and $I_{\neg\text{top-k}}$ separately, we simplify the learning process by establishing a single contrastive objective to estimate $\Delta I$. Specifically, we define the samples $(x, e^+)$ from the activated experts in $\mathcal{D}_{\text{top-k}}$ as positive samples. And the negative samples are sampled from the inactivated experts in $\mathcal{D}_{\neg\text{top-k}}$. The loss function of contrastive learning can be formulated as:

$$\mathcal{L}_{\text{NCE}} = -\mathbb{E}_{p(x,e^+)}\mathbb{E}_{\mathcal{D}_{\neg\text{top-k}}} \left[\log \frac{h_1\left(x, e^+\right)}{h_1(x, e^+) + \sum_{e^- \in \mathcal{D}_{\neg\text{top-k}}} h_2(x, e^-)}\right]. \tag{8}$$

Intuitively, the score function quantifies the exponential correlation between input token $x$ and expert representation $e$, assigning higher scores to positive samples and lower scores otherwise. The following theorem shows that the proposed contrastive objective can serve as an estimation for the MI gap with sufficient negative samples.

**Theorem 1** *(InfoNCE). The MI gap $\Delta I = I_{top\text{-}k}(x, e^+) - I_{\neg top\text{-}k}(x, e^-)$ can be lower bounded by the contrastive objective, as follows:*

$$\Delta I \geq \log(N) - \mathcal{L}_{\text{NCE}}, \tag{9}$$

*where $N$ is the number of negative samples from the inactivated experts.*

$I_{\text{NCE}} = \log(N) - \mathcal{L}_{\text{NCE}}$ approximates the true MI gap as $N$ increases, which is a tight lower bound. Please refer to Appendix A for full derivations.

To optimize the contrastive objective in Eq. 8, $h_1(\cdot)$ and $h_2(\cdot)$ are adopted to estimate the information density. Ideally, $h_1(\cdot)$ assigns high scores only to positive samples $(x, e^+)$, and $h_2(\cdot)$ assigns low scores only to negative samples $(x, e^-)$. Coincidentally, $(x, e^-) \sim \mathcal{D}_{\neg\text{top-k}}$ can serve as negative samples for the scoring function $h_1(\cdot)$, while $(x, e^+) \sim \mathcal{D}_{\text{top-k}}$ functions as positive samples for $h_2(\cdot)$, establishing a synergistic relationship between them. Thus, we integrate $h_1(\cdot)$ and $h_2(\cdot)$ into a single function $h(\cdot)$, which assigns high scores to $(x, e^+)$ and low scores to $(x, e^-)$. Then, we can simplify Eq. 8 into a new version:

$$\hat{\mathcal{L}}_{\text{NCE}} = -\mathbb{E}_{p(x,e^+)}\mathbb{E}_{\mathcal{D}_{\neg\text{top-k}}} \left[\log \frac{h\left(x, e^+\right)}{h(x, e^+) + \sum_{e^- \in \mathcal{D}_{\neg\text{top-k}}} h(x, e^-)}\right]. \tag{10}$$

$(x, e_{\text{top-k}})$ and $(x, e_{\neg\text{top-k}})$ form a bidirectional sample pair, with each entity acting as both a positive and negative reference to its counterpart. For implementation, we choose the exponential similarity function:

$$h(x, e) = \exp(E^+(x) \cdot e)/\tau, \tag{11}$$

which is commonly used to measure the similarity between two representations. Here, $\tau$ is a temperature hyperparameter.

In optimization, the score function assigns high scores to the representations of activated experts and low scores to inactivated experts. The derived InfoNCE loss in Eq. 10 can be generalized as a common contrastive loss: for each query $q_i = E_i^+(x)$, its positive keys $k^+$ are obtained from the representations of other activated experts in $k^+ \sim \mathcal{D}_{\text{top-k}}$, $k^+ \neq q_i$, while negative keys $k^-$ are sampled from $k^- \sim \mathcal{D}_{\neg\text{top-k}}$. The contrastive loss can be written as:

$$\mathcal{L}_{\text{con}} = \sum_{i=1}^{k} - \log \left(\frac{\exp(q_i \cdot k_i^+/\tau)}{\exp(q_i \cdot k_i^+/\tau) + \sum_{k_i^-} \exp(q_i \cdot k_i^-/\tau)}\right), \tag{12}$$

where $k$ denotes the number of activated experts.

| Method | Params | ST/MT | ARC-e | ARC-c | BoolQ | OBQA | PIQA | Avg. |
|---|---|---|---|---|---|---|---|---|
| LoRA | 2.9% | ST | 73.8 | 50.9 | 62.2 | 80.4 | 82.1 | 69.9 |
| | | MT | 61.3 ( -12.5 ) | 55.7 ( +4.8 ) | 66.7 ( +4.5 ) | 71.6 ( -8.8 ) | 72.4 ( -9.7 ) | 65.5 ( -4.4 ) |
| DoRA | 2.9% | ST | 76.5 | 59.8 | 71.7 | 80.6 | 82.7 | 74.3 |
| | | MT | 64.5 ( -12 ) | 54.1 ( -5.7 ) | 65.4 ( -6.3 ) | 75.8 ( -4.8 ) | 71.9 ( -10.8 ) | 66.3 ( -8 ) |
| MOELoRA | 1.0% | ST | 76.8 | 60.2 | 72.0 | 81.1 | 82.7 | 74.6 |
| | | MT | 76.1 ( -0.7 ) | 59.3 ( -0.9 ) | 71.5 ( +0.1 ) | 80.7 ( -0.4 ) | 82.1 ( -0.3 ) | 73.9 ( -0.5 ) |
| MiLoRA | 0.93% | ST | 77.8 | 61.2 | 72.8 | 81.7 | 83.3 | 75.4 |
| | | MT | 77.4 ( -0.4 ) | 61.5 ( +0.3 ) | 72.3 ( -0.3 ) | 81.3 ( -0.4 ) | 83.5 ( +0.3 ) | 75.2 ( -0.1 ) |
| MixLoRA | 2.9% | ST | 78.4 | 56.1 | 72.7 | 81.6 | 83.2 | 74.4 |
| | | MT | 76.6 ( -1.8 ) | 64.2 ( +8.1 ) | 71.2 ( -1.5 ) | 81.6 ( -0.0 ) | 82.7 ( -0.5 ) | 75.3 ( +0.9 ) |
| OMoE-LoRA | 0.73% | ST | 79.3 | 56.6 | 73.5 | 80.6 | 84.5 | 74.9 |
| | | MT | 79.8 ( +0.5 ) | 66.8 ( +10.2 ) | 72.4 ( -1.1 ) | 76.8 ( -3.8 ) | 81.6 ( -2.9 ) | 75.4 ( +0.5 ) |
| CoMoE-LoRA (ours) | 1.45% | ST | 80.3 | 57.3 | 72.9 | 80.4 | 83.6 | 74.9 |
| | | MT | 79.6 ( -0.7 ) | 66.5 ( +9.2 ) | 71.8 ( -1.1 ) | 81.2 ( +0.8 ) | 81.8 ( -1.8 ) | **76.2** ( +1.3 ) |

Table 1: Overall comparison of different PEFT methods for multi-task learning. The backbone model is LLaMA-2 7B. ST refers to the single-task settings, while MT refers to the multi-task settings. Reported results are accuracy scores, with differences between MT and ST indicated in red for decreases and in blue for increases.

| Method | Params | ARC-e | ARC-c | BoolQ | OBQA | PIQA | SIQA | HellaS | WinoG |
|---|---|---|---|---|---|---|---|---|---|
| LoRA | 2.9% | 73.8 | 50.9 | 62.2 | 80.4 | 82.1 | 69.9 | 88.4 | 66.8 |
| DoRA | 2.9% | 76.5 | **59.8** | 71.7 | 80.6 | 82.7 | 74.1 | 89.6 | 67.3 |
| MixLoRA | 2.9% | 78.4 | 56.1 | 72.7 | **81.6** | 83.2 | 78.0 | 92.8 | 76.8 |
| **CoMoE-LoRA** | 1.45% | **80.3** | 57.3 | 72.9 | 80.4 | 83.6 | **79.2** | **93.2** | **77.3** |
| **CoMoE-DoRA** | 1.45% | 80.2 | 57.0 | **73.3** | 81.2 | **83.8** | 79.1 | 92.8 | **77.3** |

Table 2: Overall comparison of different PEFT methods for single-task learning, using base models with different numbers of parameters. Bold indicates the best results.

### 4.4 Training Approach

In standard supervised fine-tuning, the primary objective is to minimize the cross-entropy loss $\mathcal{L}_{\text{CE}}$ between predicted tokens and target tokens.

As a core component in MoE, the top-$k$ router directly determines which experts are activated during inference. Based on our analysis and derivations in Sections 4.2 and 4.3, we incorporate the contrastive loss as an auxiliary optimization objective, with the complete workflow illustrated in Fig. 2. The contrastive loss effectively enhances the distinctiveness of experts, thereby promoting specialization and reducing redundancy among the experts. By incorporating the contrastive loss in Eq. 12, the total loss is computed as:

$$\mathcal{L}_{\text{total}} = \mathcal{L}_{CE} + \lambda \cdot \mathcal{L}_{\text{con}}, \tag{13}$$

where $\lambda$ is a hyperparameter to scale the auxiliary loss. Our method requires no pretraining and can be seamlessly integrated into existing MoE architectures. To summarize, the complete algorithm pseudocode is provided in Appendix B.

## 5 Experiments

In this section, we conduct extensive experiments coupled with ablation and visualization experiments to evaluate the effectiveness of CoMoE, accompanied by concise analyses.

### 5.1 Experimental Settings

**Datasets and Benchmarks.** We conduct experiments on a collection of tasks: (a) Diverse commonsense reasoning datasets: ARC-e and ARC-c (Clark et al., 2018), OpenBookQA (OBQA) (Mihaylov et al., 2018), PIQA (Bisk et al., 2020), SocialIQA (SIQA) (Sap et al., 2019), and BoolQ (Clark et al., 2019). (b) A science completion task: Hellaswag (Zellers et al., 2019). (c) A fill-in-the-blank task: Winogrande (Sakaguchi et al., 2021). (d) Math reasoning tasks: GSM8K (Cobbe et al., 2021) (grade-school word problems), AddSub (Hosseini et al., 2014) (single-step arithmetic), AQuA (Ling et al., 2017) (algebra multiple-choice), MultiArith (Roy and Roth, 2015) (multi-step arithmetic), SingleEQ (Koncel-Kedziorski et al., 2015) (single-equation problems), and SVAMP (Patel et al., 2021) (robustness testing). (e) Symbolic math expression evaluation: Arithmetic (Brown et al., 2020).

We utilize the PEFT framework provided by (Hu et al., 2023; Li et al., 2024) for training on these datasets. We choose LLaMA-2 7B and Gemma 2B as our backbone models. The detailed statistics and evaluation metrics can be found in Appendix C.

| Method | GSM8K | AddSub | AQuA | MultiArith | SingleEQ | SVAMP | Arithmetic | Avg. |
|---|---|---|---|---|---|---|---|---|
| LoRA | 31.1 | 84.8 | 17.6 | 88.3 | 90.2 | 65.0 | 2.69 | 54.2 |
| CoMoE-LoRA | 32.2 | 89.9 | 31.4 | 91.7 | 90.2 | 66.0 | 80.9 | 68.90 |

Table 3: Performance comparison between LoRA and CoMoE-LoRA on math reasoning tasks. CoMoE-LoRA achieves substantial improvements under multi-task settings.

**Baselines.** In this study, we compare our method with several popular and well-established baselines to assess its performance. For multi-task settings, we evaluate CoMoE against LoRA, its variants, and MoE-based methods, including: (1) **LoRA** (Hu et al., 2022a); (2) **DoRA** (Liu et al., 2024), which decomposes LoRA weights into magnitude and direction; (3) **MoELoRA** (Liu et al., 2023), which decomposes a LoRA module into a mixture of experts; (4) **MiLoRA** (Zhang et al., 2024), which treats each LoRA module as an expert and employs a routing mechanism; (5) **MixLoRA** (Li et al., 2024), a resource-efficient sparse MoE model based on LoRA; (6) **OMoE** (Feng et al., 2025), which applies hard constraints to promote diversity. For single-task settings, we compare our approach with other PEFT baselines, including: **AdaLoRA** (Zhang et al., 2023a); **Parallel-Adapter** (He et al., 2021); **Learned-Adapter** (Zhang et al., 2023b); **P-tuning v2** (Liu et al., 2021); **IAPT** (Zhu et al., 2024); **BitFit** (Ben-Zaken et al., 2022); **(IA)$^3$** (Liu et al., 2022); **SSP** (Hu et al., 2022b).

Most of the results are directly extracted from (Zhang et al., 2024), with a few baselines reproduced by running the provided source code. In single-task settings, we compare CoMoE with LoRA, DoRA, and MixLoRA as representative baselines, while comparisons with other methods are presented in Appendix E.

**Implementation Details.** To evaluate the effectiveness of CoMoE, we apply it on the basis of LoRA and DoRA, respectively, and label them as CoMoE-LoRA and CoMoE-DoRA in the experiments. Unless otherwise specified, CoMoE is configured with $r = 16$, incorporating 4 experts and a top-2 router, applied to the $q$, $k$, $v$, and $o$ parameters in the attention layers. For all settings, we adopt supervised fine-tuning only. Due to space limitations, the detailed experimental settings for baselines and hyperparameters are provided in Appendix D.

### 5.2 Main Results

**Multi-task Setup.** Table 1 summarizes the multi-task performance of CoMoE and baselines on LLaMA-2 7B. The results verify that both LoRA and DoRA, due to their lack of MoE structures, struggle to handle heterogeneous datasets, observing a significant drop in multi-task settings (7% $\sim$ 12% degradation). MoE-based PEFT methods (MoELoRA, MiLoRA, MixLoRA, and OMoE) mitigate the performance degradation in multi-task settings but fail to leverage the modularity and specialization in MoE, leaving potential performance gains unexploited. In contrast, CoMoE introduces contrastive learning to promote the modularization and specialization of experts, enabling efficient utilization of the MoE's capacity. Thus, CoMoE not only improves parameter efficiency (reduces the number of experts) but also achieves an average accuracy gain of +1.3. In addition, we evaluate CoMoE-LoRA on several math reasoning tasks under more complex multi-task settings, as shown in Table 3. The results related to LoRA are reproduced from (Luo et al., 2024). Compared with LoRA, CoMoE achieves a notable improvement in performance, further demonstrating its generalization in math reasoning tasks.

**Single-task Setup.** In this setup, we compare the performance of CoMoE and baselines in single-task settings. The experimental results are shown in Table 2. Comparisons with other baselines are provided in Appendix E. CoMoE demonstrates superior parameter efficiency while maintaining comparable performance, with a reduction of approximately $50\%$ in tunable parameters. Remarkably, the diversity among experts even yields accuracy improvements on a subset of datasets (e.g., ARC-e, BoolQ, and SIQA). The performance differences between multi-task and single-task settings are further analyzed in Appendix J.

### 5.3 Ablation Studies and Further Analysis

**Effects of the Hyperparameter $\lambda$.** We further evaluate $\lambda \in \{0.0, 0.001, 0.01, 0.1, 1.0\}$ on the BoolQ and the multi-task settings (ARC-c, ARC-e, BoolQ, and OBQA). Experimental results in Fig. 3(a) and (b) demonstrate that $\lambda = 1e{-}2$ yields optimal performance in both single-task and multi-task settings. In single-task configurations, model performance shows gradual improvement as the value increases from 0, 0.001 to 0.01, suggesting that even individual tasks can benefit from diversity. However, significant performance degradation occurs at higher values (0.1 and 1.0), indicating that excessive diversity impedes effective dataset adaptation. This phenomenon is more pronounced in multi-task settings, where the model shows amplified gains from diversity.

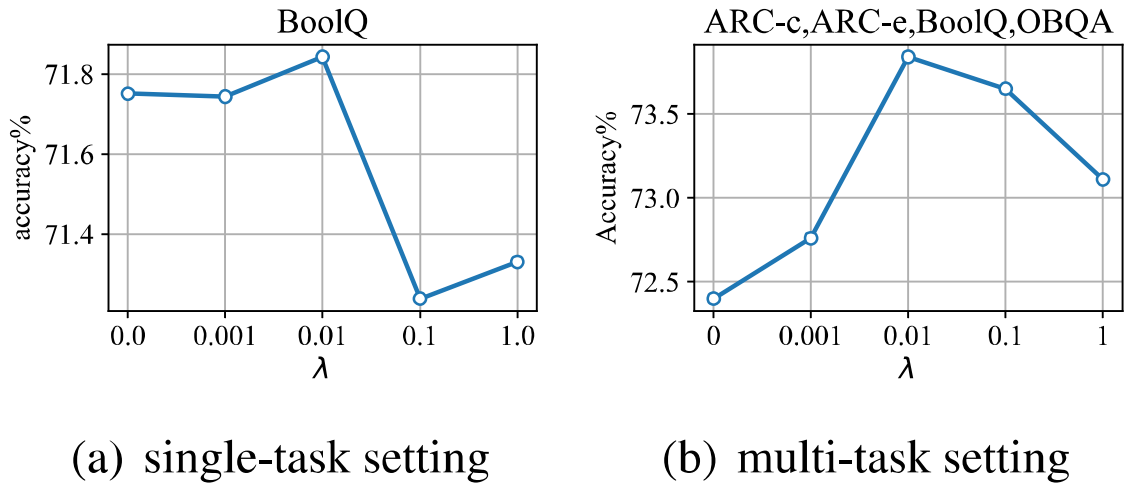

(a) single-task setting (b) multi-task setting

Figure 3: Performances under different coefficients $\lambda$. (a) In single-task settings. (b) In multi-task settings.

| Method | ARC-e | ARC-c | BoolQ | OBQA | Avg. |
|---|---|---|---|---|---|
| CoMoE-$\triangle$ | 80.5 | 67.0 | 72.4 | 78.4 | **74.6** |
| CoMoE-$\triangledown$ | 79.7 | 65.1 | 72.3 | 76.8 | **73.5** |
| CoMoE-$\Diamond$ | 77.9 | 65.3 | 72.1 | 77.6 | **73.2** |
| CoMoE-$\bowtie$ | 77.7 | 66.6 | 72.4 | 80.0 | **74.2** |

Table 4: Performance comparison of different CoMoE variants on four datasets.

**Effects of the Number of Experts $n$.** The results are provided in Appendix F. We can see that CoMoE benefits from more experts.

**Ablation on Different Backbones.** We conduct multi-task experiments on Gemma 2B, with detailed results provided in Appendix G.

**Model Efficiency.** Please refer to Appendix H.

**Layer-wise Diversity Analysis.** Having established the benefit of expert diversity in multi-task settings, we naturally ask: Which layers in LLMs benefit most from diverse experts? Using LLaMA-2 7B as a case, we simplify the large language model into three levels: low (from layer 1 to 10), medium (from layer 11 to 20), and high (from layer 21 to 32), and inject CoMoE into them. Specifically, four types of layer-wise diversity configurations are explored: (1) **CoMoE-$\triangle$**, applying CoMoE in low layers; (2) **CoMoE-$\triangledown$**, applying CoMoE in high layers; (3) **CoMoE-$\Diamond$**, applying CoMoE in medium layers; (4) **CoMoE-$\bowtie$**, applying CoMoE in low and high layers. Our experiments in Table 4 reveal that CoMoE-$\triangle$ and CoMoE-$\bowtie$ achieve superior performance, outperforming CoMoE-$\triangledown$ and CoMoE-$\Diamond$ by 0.7$\sim$1.4 on average accuracy across datasets. This performance gap stems from the placement of CoMoE in lower transformer layers, which is critical for establishing diversity early in the processing hierarchy.

**Analysis of Workload Balance.** The primary distinction between CoMoE and vanilla MoE variants lies in the contrastive learning objective that promotes specialization and modularity among experts. To further elucidate the effectiveness of CoMoE, we conducted an in-depth analysis of expert activation in MoE under a multi-task setting. Fig. 4 presents a comparative visualization of expert workload distributions before and after introducing the contrastive loss. Without contrastive loss, all tasks predominantly concentrated on expert 1 and 2, indicating insufficient differentiation and an imbalanced workload. After introducing contrastive loss, distinct tasks exhibited marked differences in expert selection. For ARC-c, expert 1 and 3 showed significantly increased activation frequency. For BoolQ, expert 1 and 4 formed a stable collaborative relationship. Notably, our method did not incorporate any routing balance loss, yet the collaboration among experts emerged naturally.

**Visualization of Expert Representations.** To demonstrate the impact of contrastive loss on expert diversity, we conduct qualitative analysis by visualizing expert representations before and after its incorporation, which is illustrated in Fig. 5. In multi-task settings, we choose the OBQA dataset for example and the visualizations of other datasets are provided in Appendix I. The results demonstrate that the absence of contrastive loss leads to substantial redundancy among experts, compromising their discriminability. In contrast, the introduction of contrastive loss fosters divergent experts, culminating in specialization and modularity.

## 6 Conclusion

In this paper, we focus on the problem of redundant knowledge learning in MoE, which leads to the underutilization of its capacity. To address this issue, we propose Contrastive Representation for MoE (CoMoE), a novel MoE variant that promotes modularization and specialization. Specifically, we define a mutual information (MI) gap between activated and inactivated experts and approximate it through a contrastive objective. This objective effectively captures the MI gap and is incorporated into supervised fine-tuning as an auxiliary optimization term. Experiments on various tasks demonstrate that CoMoE outperforms the baselines in multi-task settings and enhances expert modularity.

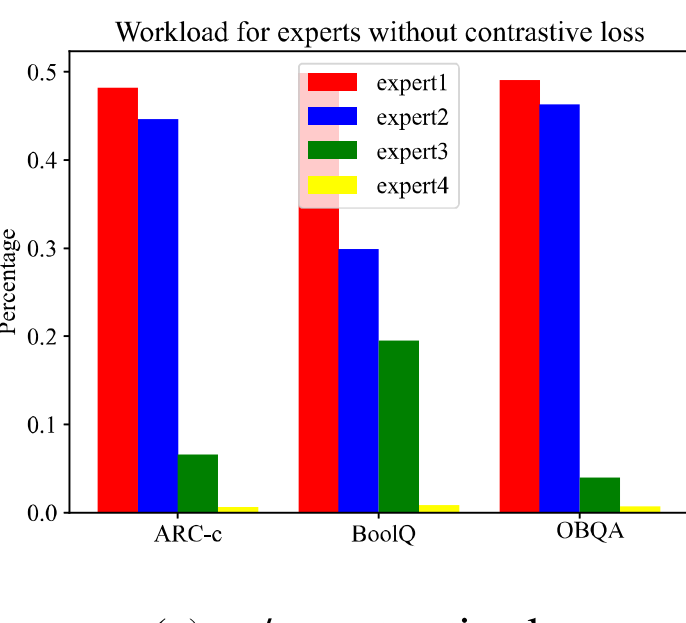

(a) w/o contrastive loss

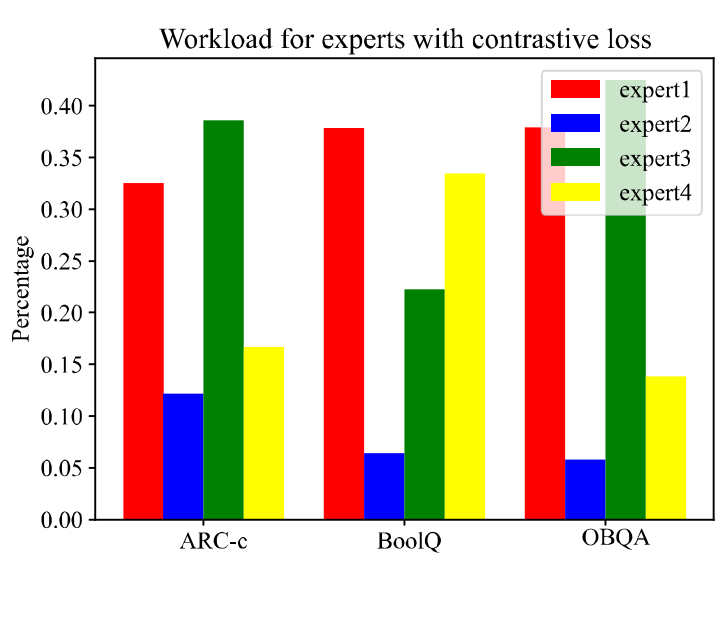

(b) w/ contrastive loss

Figure 4: Comparison of the workloads of experts before and after contrastive loss incorporation in a multi-task setting. (a) Without contrastive loss. (b) With contrastive loss.

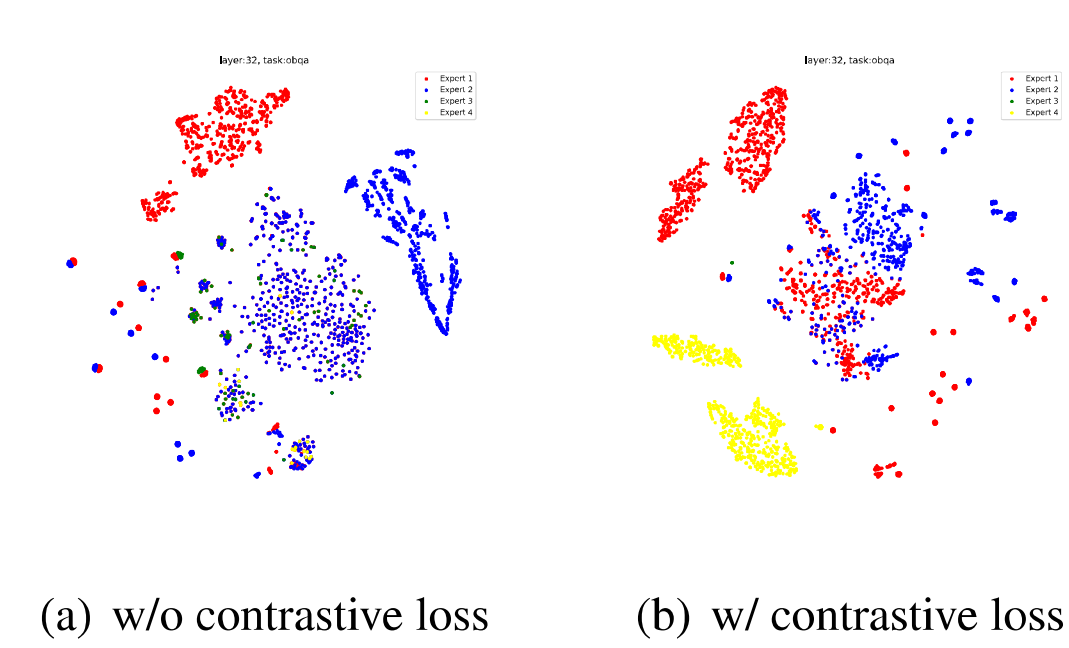

(a) w/o contrastive loss (b) w/ contrastive loss

Figure 5: Comparison of expert representations in obqa before and after contrastive loss incorporation in a multi-task setting. (a) Without contrastive loss. (b) With contrastive loss.

## Limitations

We show that our proposed method can significantly improve the specialization of experts and the performance of MoE in multi-task settings. However, there are several limitations to acknowledge: (a) The computational cost of CoMoE. CoMoE leverages inactivated experts as negative samples, and its computational cost is proportional to $O(N)$. To alleviate the computational burden, a complexity reduction strategy and empirical validation are provided in Appendix H. However, it substantially improves expert specialization and modularization, achieving remarkable performance with a limited number of experts; (b) Other evaluation benchmarks were not considered. Nevertheless, CoMoE is simple and can be easily integrated with different backbone models and various downstream tasks.

## Ethics Statement

This paper proposes a novel method to enhance parameter-efficient fine-tuning based on the Mixture-of-Experts architecture, which simultaneously improves expert modularization and specialization while boosting performance on heterogeneous and complex datasets. The experiments employ widely adopted benchmark datasets in the research community that, to our knowledge, involve no privacy concerns or ethical controversies. The experiments are conducted on the open-source LLaMA-2 series of large language models. It should be emphasized that this work represents fundamental research focused exclusively on advancing MoE-based fine-tuning methods for LLMs, rather than developing applications.

## Acknowledgments

This work was supported in part by the Key Research Project of Chinese Academy of Sciences under Grant No. KGFZD-145-25-21, the National Natural Science Foundation of China under Grant 62322316, and the Open Fund of National Key Laboratory of Information Systems Engineering (No. 6142101240203).

## A Theoretical Proof

**Theorem 2** *(InfoNCE). The MI gap* $\Delta I = I_{top\text{-}k}(x, e^+) - I_{\neg top\text{-}k}(x, e^-)$ *can be lower bounded by the contrastive objective, as*

$$\Delta I \geq \log(N) - \mathcal{L}_{\text{NCE}}, \tag{14}$$

*where* $N$ *is the number of negative samples from the inactivated expert dataset.*

*Proof.* As illustrated in Eq. 8, we utilize $h_1(x, e^+) \propto p(e^+|x)/p(e^+)$ and $h_2(x, e^-) \propto p(e^-|x)/p(e^-)$ to approximate the information density ratio, which preserves the MI between $x$ and $M$. Based on the standard derivations presented in InfoNCE (Oord et al., 2018) and (Wen et al., 2024), the contrastive objective can be rewritten as:

$$\begin{aligned}
\mathcal{L}_{\text{NCE}} &= -\mathbb{E}_{p(x,e^+)}\mathbb{E}_{\mathcal{D}_{\neg\text{top-k}}} log \left[ \frac{p(e^+|x)/p(e+)}{p(e^+|x)/p(e^+) + \sum_{e^- \in \mathcal{D}_{\neg\text{top-k}}} p(e^-|x)/p(e^-)} \right] \\
&= \mathbb{E}_{p(x,e^+)}\mathbb{E}_{\mathcal{D}_{\neg\text{top-k}}} log \left[ 1 + \frac{p(e^+)}{p(e^+|x)} \sum_{e^- \in \mathcal{D}_{\neg\text{top-k}}} \frac{p(e^-|x)}{p(e^-)} \right] \\
&= \mathbb{E}_{p(x,e^+)}\mathbb{E}_{\mathcal{D}_{\neg\text{top-k}}} log \left[ 1 + N \frac{p(e^+)}{p(e^+|x)} (\frac{1}{N}) \sum_{e^- \in \mathcal{D}_{\neg\text{top-k}}} \frac{p(e^-|x)}{p(e^-)} \right] \\
&\geq \mathbb{E}_{p(x,e^+)} log \left[ N \frac{p(e^+)}{p(e^+|x)} (\frac{1}{N}) \sum_{e^- \in \mathcal{D}_{\neg\text{top-k}}} \frac{p(e^-|x)}{p(e^-)} \right] \\
&= \mathbb{E}_{p(x,e^+)} log \left[ (\frac{1}{N}) \sum_{e^- \in \mathcal{D}_{\neg\text{top-k}}} N \frac{p(e^+)}{p(e^+|x)} \frac{p(e^-|x)}{p(e^-)} \right]
\end{aligned} \tag{15}$$

We derive Eq. 15 using Jensen's inequality, noting that log is a concave function.

$$\begin{aligned}
\mathcal{L}_{\text{NCE}} &\geq \mathbb{E}_{p(x,e^+)} \left[ (\frac{1}{N}) \sum_{e^- \in \mathcal{D}_{\neg\text{top-k}}} log \left[ N \frac{p(e^+)}{p(e^+|x)} \frac{p(e^-|x)}{p(e^-)} \right] \right] \\
&= \mathbb{E}_{p(x,e^+)} \left[ (\frac{1}{N}) \sum_{e^- \in \mathcal{D}_{\neg\text{top-k}}} \left[ logN + log \frac{p(e^+)}{p(e^+|x)} + log \frac{p(e^-|x)}{p(e^-)} \right] \right] \\
&\approx \mathbb{E}_{p(x,e^+)} \left[ logN + log \frac{p(e^+)}{p(e^+|x)} + \mathbb{E}_{e^- \in \mathcal{D}_{\neg\text{top-k}}} log \frac{p(e^-|x)}{p(e^-)} \right] \\
&= logN - I_{\text{top-k}} + I_{\neg\text{top-k}} \\
&= lonN - \Delta I
\end{aligned} \tag{16}$$

Thus, we prove the aforementioned theorem:

$$\Delta I \geq \log(N) - \mathcal{L}_{\text{NCE}}. \tag{17}$$

## B Optimization Algorithm

To enhance specialization and modularization among experts, we devise *Contrastive Representation for MoE*, outlined in Algorithm 1. For each sample, one expert is randomly selected from its top-$k$ expert indices $T$ as the anchor. The remaining top-$k$ experts act as positives, while all other inactivated experts form negatives. The procedure is:

1. **Expert representation.** Obtain the individual expert outputs $\{E_j(x)\}_{j=1}^n$, each with dimensionality $E_j(x) \in \mathbb{R}^D$.

**Algorithm 1:** Contrastive Loss Computation among Experts (single-sample)

**Input:** Top-$k$ expert indices $T \in \mathbb{N}^k$
Expert representations $\{E_j(x)\}_{j=1}^{n}$, where $E_j(x) \in \mathbb{R}^D$
Temperature $\tau$
**Output:** Contrastive loss $\mathcal{L}_{\text{contrast}}$

```
1  r ~ U{1, ..., k}                                   // Random anchor position
2  a ← T[r]                                           // Anchor expert index
3  q ← Normalize(E_a(x))
4  P ← {Normalize(E_T[j](x)) | j ≠ r}                 // Positive set, size k − 1
5  N ← {Normalize(E_j(x)) | j ∉ T}                    // Negative set, size n − k
6  s_pos ← (q · P^T)/τ
7  s_neg ← (q · N^T)/τ
8  logits ← [ s_pos, s_neg ]
9  L_contrast = −log( Σ exp(s_pos) / (Σ exp(logits) + ε) )
10 return L_contrast
```

2. **Anchor selection.** Uniformly sample an index $r$ from $\{1, \ldots, k\}$. Define the anchor (query) vector as $q = \text{Normalize}(E_{T[r]}(x))$.

3. **Positive set.** Aggregate the remaining $(k - 1)$ expert representations, excluding the one indexed by $T[r]$ from the index set $T$, into a set $P$, applying normalization:

$$P = \{\text{Normalize}(E_{T[j]}(x)) \mid j \neq r\}.$$

4. **Negative set.** Collect and normalize representations from experts not included in the top-$k$ indices:

$$N = \{\text{Normalize}(E_j(x)) \mid j \notin T\}.$$

5. **Similarity computation.** Compute cosine similarities between anchor vector $q$ and each representation in $P$ and $N$, scaled by temperature $\tau$, yielding similarity scores $s_{\text{pos}}$ and $s_{\text{neg}}$:

$$s_{\text{pos}} = \frac{q \cdot P^{\top}}{\tau}, \quad s_{\text{neg}} = \frac{q \cdot N^{\top}}{\tau}.$$

6. **InfoNCE loss.** Concatenate the logits and compute the InfoNCE loss as

$$\mathcal{L}_{\text{contrast}} = -\log\left(\frac{\sum \exp(s_{\text{pos}})}{\sum \exp([s_{\text{pos}}, s_{\text{neg}}]) + \varepsilon}\right).$$

where $\varepsilon$ is a small positive value (e.g., $10^{-3}$) used to ensure numerical stability, avoiding computational issues caused by the denominator being zero.

## C Datasets

Detailed information about the datasets used in the experiments is presented in Table 5. All datasets are downloaded from HuggingFace.

## D Experimental settings

**Computing Infrastructure** We run all our experiments on NVIDIA A6000 (48GB) GPUs, using Python 3.10 and Ubuntu 20.04 on x86-64 CPUs.

**Pretrained Backbones** The main experiments use the most recent open-sourced LLM, LLaMA-2 7B and Gemma 2B, as the pretrained backbone model. When fine-tuning LLaMA-2 7B and Gemma 2B, we consider only the supervised fine-tuning setting.

**Hyperparameters for CoMoE** In our experiments, unless otherwise specified, we set the hyperparameters as illustrated in Table 6. In the table, the hyperparameters set by other baseline methods, LoRA, DoRA, MixLoRA, MixDoRA, OMoE-LoRA, are also included. Under this setting, CoMoE introduces approximately 1.45% tunable parameters to the LLaMA-2 7B backbone.

**Descriptive Statistics about Results** We conduct experiments on all training settings using five different random seeds, and the final results represent the median accuracy within each setting.

## E Additional Results on Other Baselines

In the main paper, we compared CoMoE with three widely recognized and well-performing baselines (LoRA, DoRA, and MixLoRA) using the LLaMA-2 7B model. In addition to the results shown in Table 2, we provide experimental results involving 11 additional strong baselines on the same LLaMA-2 7B backbone, as detailed in Table 12. The results demonstrate that CoMoE achieves significant improvements in both parameter efficiency and overall performance compared to these baselines.

| Datasets | #train | #test | Type | Metrics |
|---|---|---|---|---|
| BoolQ | 9,427 | 3,270 | Text Classification | acc |
| OBQA | 4,957 | 500 | Question Answering | acc |
| ARC-e | 2,251 | 2,376 | Question Answering | acc |
| ARC-c | 1,119 | 1,172 | Question Answering | acc |
| PIQA | 16,100 | 1,840 | Question Answering | acc |
| SIQA | 33,410 | 1,954 | Question Answering | acc |
| HellaSwag | 39,905 | 10,042 | Sentence Completion | acc |
| WinoGrande | 9,248 | 1,267 | Fill in the Blank | acc |

Table 5: The dataset statistics.

| **Hyperparameters** | **LoRA/DoRA** | **MixLoRA/MixDoRA** | **OMoE-LoRA** | **CoMoE** |
|---|---|---|---|---|
| Cutoff Length | 512 | | | |
| Learning Rate | 2e-4 | | | |
| Optimizer | AdamW | | | |
| Batch size | 16 | | | |
| Accumulation Steps | 8 | | | |
| Dropout | 0.05 | | | |
| Epochs | 2 | | | |
| Where | Q, K, V, O, Up, Down, Gate | | | |
| LoRA Rank $r$ | 80 | 16 | 16 | 16 |
| LoRA Alpha $\alpha$ | 160 | 32 | 32 | 32 |
| Experts | - | 8 | 2 | 4 |
| Routing strategy | - | Top - 2 routing | Soft routing | Top - 2 routing |

Table 6: Hyperparameter configurations of LoRA, DoRA, MixLoRA, MixDoRA, OMoE-LoRA and CoMoE for fine-tuning LLaMA-2 7B on datasets.

## F Additional Results on Different Number of Experts $n$

We further compared the experimental results under different numbers of experts. In the multi-task setting, using the ARC-c, ARC-e, BoolQ, and OBQA datasets, the results are shown in Table 7. As the number of experts increases from 4 to 8, the model's performance stabilizes and shows modest improvement. While merely increasing the number of experts does not guarantee significant performance enhancement, expanding the expert pool in multi-task settings contributes to improved model stability and may lead to certain performance gains.

## G Additional Results on Different Backbone Models

Our main experiments are conducted on LLaMA-2 7B. To demonstrate the adaptation of our method, we compare CoMoE with MixLoRA on the Gemma 2B backbone, as shown in Table 8. Results indicate that CoMoE still achieves certain performance gains over the baseline across different backbones.

## H Model efficiency and Computational budget

To evaluate the model efficiency and computational budget of CoMoE, we compare it with the baselines(LoRA, DoRA, MixLoRA, OMOE) in three aspects: inference latency, memory cost, and training time. We base our evaluation on the following three metrics: (a) the inference time required for generating responses (ms), (b) the GPU memory cost (MiB), and (c) the training time in multi-task settings (h). In the multi-task setting, ARC-e, ARC-c, BoolQ, and OBQA are trained simultaneously. The results are provided in Table 9. From Table 9, we observe that compared to the well-performing MixLoRA, CoMoE achieves a 10% improvement in inference efficiency while reducing GPU memory usage by 465 MiB. In terms of training time, CoMoE requires 3.5 hours of training on an A6000 GPU under the multi-task setting. Although CoMoE increases the training burden, it does not compromise inference efficiency and simultaneously enhances model performance. Furthermore, we conducted experiments to investigate the relationship between training time and the number of experts, as shown in Table 10. In our proposed method, negative samples are drawn from the set of inactive experts. As a result, the training cost increases linearly with the number of experts, with a time complexity of $O(n)$. To mitigate this computational overhead, we introduce a fixed-size sampling strategy, wherein a constant number of negative samples are randomly selected from the inactive experts at each training step. This reduces the complexity from $O(n)$ to constant time $O(1)$, rendering it independent of the total number of experts. The results of this strategy are summarized in Table 11. Theoretically, this approach introduces a looser lower bound (see Theorem 1); however, empirical results show no degradation in performance, as evidenced by the comparable average accuracy between Table 10 and Table 11.

| Number of Experts $n$ | ARC-e | ARC-c | BoolQ | OBQA | Avg. |
|---|---|---|---|---|---|
| $n = 4$ | 80.0 | 66.6 | 71.2 | 77.6 | 73.9 |
| $n = 5$ | 79.8 | 64.4 | 70.9 | 77.4 | 73.1 |
| $n = 6$ | 80.6 | 64.5 | 73.1 | 76.0 | 73.6 |
| $n = 7$ | 79.7 | 65.4 | 72.5 | 78.0 | 73.9 |
| $n = 8$ | 79.1 | 64.7 | 72.0 | 80.8 | 74.2 |

Table 7: Accuracy results across different expert configurations (from 4 to 8 experts) on multi-task evaluation.

| Method | ARC-e | ARC-c | BoolQ | OBQA | Avg. |
|---|---|---|---|---|---|
| MixLoRA | 22.4 | 24.0 | 62.2 | 27.6 | **34.1** |
| CoMoE-LoRA | 25.7 | 23.7 | 62.2 | 25.2 | **34.2** |

Table 8: Comparison of MixLoRA and CoMoE in multi-task learning. The backbone model is Gemma 2B.

| Method | Latency (ms) | Memory (MiB) | Training time (h) |
|---|---|---|---|
| LoRA | 2,096 | +1,630 | 1.8h |
| DoRA | 1,748 | +2,184 | 1.7h |
| MixLoRA | 4,217 | +1,776 | 2.2h |
| OMoE(Top-2) | 4,863 | +1,776 | 2.3h |
| CoMoE | 3,789 | +1,311 | 3.5h |

Table 9: The inference latency, memory cost and training time of the LLaMA-2 7B for generating a batch of responses using CoMoE and baselines.

## I Additional Visualization of Representations under Different Datasets

Our main experiment visualizes the expert representations of the obqa dataset in multi-task settings before and after introducing the contrastive loss, which is illustrated in Fig. 5. The visualization results demonstrate that the introduced contrastive loss promotes modularity among experts while preventing knowledge redundancy between them. Visualization results from other datasets (ARC-c and BoolQ) under the same model, which are provided in Fig. 6 and Fig. 7, yield similar conclusions.

## J Analysis of Performance Differences Between Multi-task and Single-task Settings

CoMoE exhibits consistently stronger performance in multi-task setups compared to single-task training, as demonstrated in both math and reasoning benchmarks. This observation suggests that the expert specialization mechanism of CoMoE is particularly well-suited for heterogeneous data distributions. In contrast, under simple or narrow single-task settings (e.g., ARC-c or OBQA), models may overfit to the specific task distribution and potentially achieve better results through memorization rather than generalization. To further illustrate this difference, consider the following math problem:

> *"Eight more than four times the number of coffee mugs in the teacher's lounge is 40. How many coffee mugs are in the teacher's lounge?"*

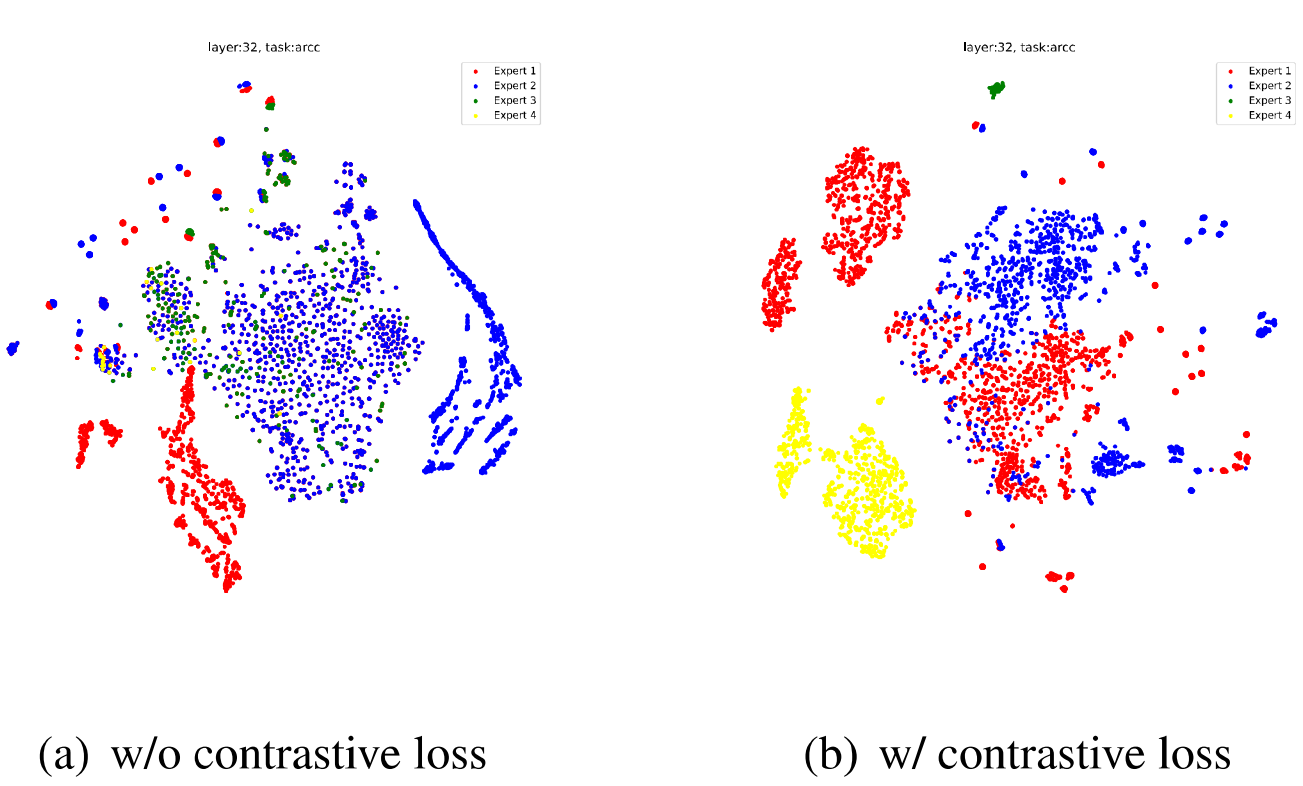

(a) w/o contrastive loss (b) w/ contrastive loss

Figure 6: Comparison of expert representations in ARC-c before and after contrastive loss incorporation in a multi-task setting. (a) Without contrastive loss. (b) With contrastive loss.

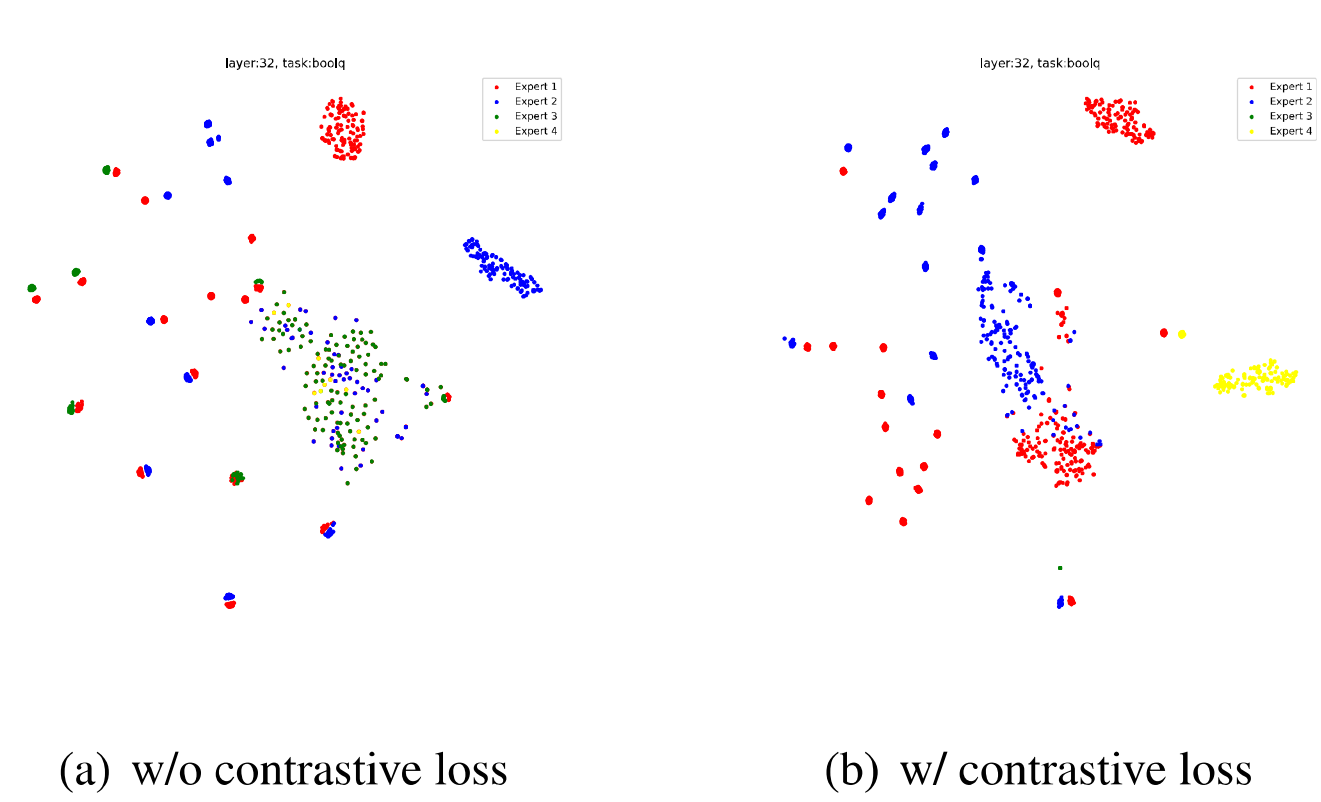

(a) w/o contrastive loss (b) w/ contrastive loss

Figure 7: Comparison of expert representations in BoolQ before and after contrastive loss incorporation in a multi-task setting. (a) Without contrastive loss. (b) With contrastive loss.

| Number of Experts | 4 | 5 | 6 | 7 | 8 |
|---|---|---|---|---|---|
| **Training Time (h)** | 7.5 | 9.0 | 10.8 | 12.0 | 13.0 |
| **Average Accuracy** | 0.7248 | 0.7309 | 0.7248 | 0.7381 | 0.7239 |

Table 10: Training time and accuracy with varying number of experts. Training time grows linearly with expert count.

| Number of Experts | 4 | 5 | 6 | 7 | 8 |
|---|---|---|---|---|---|
| **Training Time (h)** | 7.5 | 7.5 | 7.5 | 7.6 | 7.6 |
| **Average Accuracy** | 0.7424 | 0.7357 | 0.7311 | 0.7305 | 0.7319 |

Table 11: Training time and accuracy after applying fixed-size negative sampling. Training time remains constant regardless of expert count.

| Method | Params | ARC-e | ARC-c | BoolQ | OBQA | PIQA |
|---|---|---|---|---|---|---|
| ***Baselines*** | | | | | | |
| Parallel-Adapter | 0.96% | 67.1 | 54.2 | 65.2 | 76.3 | 69.8 |
| Learned-Adapter | 0.94% | 69.3 | 54.4 | 64.9 | 78.4 | 75.6 |
| P-tuning v2 | 0.97% | 63.5 | 51.3 | 61.2 | 76.1 | 66.2 |
| IAPT | 0.96% | 66.3 | 54.7 | 67.8 | 79.2 | 77.3 |
| BitFit | 1.00% | 65.9 | 54.1 | 66.4 | 77.2 | 76.6 |
| $(IA)^3$ | 0.90% | 68.1 | 54.6 | 67.2 | 78.1 | 75.4 |
| SSP | 0.93% | 71.6 | 57.6 | 69.6 | 79.5 | 79.7 |
| AdaLoRA | 0.92% | 73.8 | 57.9 | 69.2 | 80.4 | 82.1 |
| MOELoRA | 1.00% | 76.8 | 60.2 | 72.0 | 81.1 | 82.7 |
| MiLoRA | 0.93% | 77.8 | 61.2 | 72.8 | 81.7 | 83.3 |
| MiDoRA | 0.93% | 77.5 | 61.3 | 72.9 | 81.3 | 83.1 |
| ***Our proposed methods*** | | | | | | |
| CoMoE-LoRA | 1.45% | **80.3** | 57.3 | 72.9 | 80.4 | 83.6 |
| CoMoE-DoRA | 1.45% | 80.2 | 57.0 | **73.3** | 81.2 | **83.8** |

Table 12: An overall comparison of different PEFT methods for single-task learning. Most results are extracted from the original papers or reproduced by running the provided source code. The backbone model is LLaMA-2 7B.

A model trained under the single-task setting responds:

> *"Assistant: Eight more than four times the number of coffee mugs in the teacher's lounge is 40. That means four times the number of coffee mugs in the teacher's lounge is 40 – 8 = 32. So, the number of coffee mugs in the teacher's lounge is 32. The answer is 32."*

This output is incorrect—it skips formal equation modeling and instead relies on direct arithmetic subtraction, leading to an erroneous conclusion. This reflects a typical overfitting behavior, where the model learns shallow patterns from the training set rather than robust reasoning strategies. Such behavior is more likely in single-task regimes due to limited exposure to diverse reasoning formats.

In contrast, the same problem posed to a multi-task trained model yields:

> *"Assistant: Eight more than four times the number of coffee mugs is 40. We can write this as 4x + 8 = 40. We can solve this equation to find x. 4x + 8 = 40. 4x = 40 – 8. 4x = 32. x = 32 / 4 = 8. The number of coffee mugs must be 8. The answer is 8."*

This response correctly models the algebraic structure and performs a step-by-step solution. It demonstrates stronger generalization and compositional reasoning ability, which we attribute to the diversity and inductive bias introduced by multi-task training. These findings underscore the advantage of CoMoE in multi-task settings, especially for problems requiring structured reasoning.

## K An Intuitive Illustration of Expert Specialization

To clarify the intuition behind our expert specialization design, consider the following illustrative scenario: Alice, Bob, and Carol are working together to develop a robot. Initially, all three study mathematics, programming, and hardware simultaneously. However, their lack of focus leads to only moderate proficiency across all domains. Later, they adopt a collaborative specialization strategy: mathematics is primarily handled by Alice and Bob, programming by Bob and Carol, and hardware by Alice and Carol. In this arrangement, Alice takes the lead in mathematics with Bob assisting; Bob leads programming with Carol assisting; and Carol leads hardware assembly with Alice assisting.

This role assignment enables each person to develop deeper expertise in their respective areas. The result is a more effective and efficient collaboration, ultimately leading to the successful construction of the robot.

Translating this scenario into our model design, we employ contrastive learning to formalize the process of specialization. For example, we encourage Alice and Bob (experts) to focus on learning mathematical tasks while discouraging Carol's involvement in that domain. This negative signal allows Carol to concentrate on mastering programming or hardware instead. In practice, such structured encouragement and discouragement drive the emergence of distinct and complementary expert roles within the model, enabling better generalization across diverse tasks.